\documentclass{article}
\usepackage{arxiv}
\usepackage[T1]{fontenc}    
\usepackage{hyperref}       
\usepackage{url}            
\usepackage{booktabs}       
\usepackage{amsmath,amssymb,amsfonts}
\usepackage{nicefrac}       
\usepackage{microtype}      
\usepackage{lipsum}
\usepackage{graphicx}
\usepackage{multirow}
\usepackage{color}
\usepackage[most]{tcolorbox}
\usepackage{array}
\definecolor{lightgray}{gray}{0.95}
\usepackage[most]{tcolorbox}
\usepackage{tabularx}
\usepackage{enumitem}
\graphicspath{ {./images/} }
\usepackage{authblk}
\usepackage{amsmath}
\usepackage{cite}
\usepackage{amsmath,amssymb,amsfonts}
\usepackage{bbm}
\usepackage{algorithmic}
\usepackage{graphicx}
\usepackage{textcomp}
\usepackage{xcolor}
\usepackage{colortbl}
\definecolor{lightgray}{gray}{0.95}
\usepackage{booktabs}
\usepackage{makecell}
\usepackage{multirow}
\usepackage{pifont} 
\usepackage{booktabs} 
\usepackage{subfigure}
\usepackage[most]{tcolorbox}
\def\BibTeX{{\rm B\kern-.05em{\sc i\kern-.025em b}\kern-.08em
    T\kern-.1667em\lower.7ex\hbox{E}\kern-.125emX}}

\title{Membership Inference Attack against Long-Context Large Language Models}

\author[1]{Zixiong Wang}
\author[1,*]{Gaoyang Liu}
\author[2]{Yang Yang}
\author[1]{Chen Wang}

\affil[1]{Huazhong University of Science and Technology, Wuhan, China \newline zixwang@hust.edu.cn, yangyang@hubu.edu.cn, \{husterlgy, cwangwhu\}@gmail.com}
\affil[2]{Key Laboratory of Intelligent Sensing System and Security (Ministry of Education) \newline School of Artificial Intelligence, Hubei University, Wuhan, China}

\begin{document}
\maketitle 
\begin{abstract}
Recent advances in Large Language Models (LLMs) have enabled them to overcome their context window limitations, and demonstrate exceptional retrieval and reasoning capacities on longer context.
Quesion-answering systems augmented with Long-Context Language Models (LCLMs) can automatically search massive external data and incorporate it into their contexts, enabling faithful predictions and reducing issues such as hallucinations and knowledge staleness.
Existing studies targeting LCLMs mainly concentrate on addressing the so-called \textit{lost-in-the-middle} problem or improving the inference effiencicy, leaving their privacy risks largely unexplored. In this paper, we aim to bridge this gap and argue that integrating all information into the long context makes it a repository of sensitive information, which often contains private data such as medical records or personal identities. We further investigate the membership privacy within LCLM's external context, with the aim of determining whether a given document or sequence is included in the LCLM’s context. Our basic idea is that if a document lies in the context, it will exhibit a low generation loss or a high degree of semantic similarity to the contents generated by LCLMs. We for the first time propose six membership inference attack (MIA) strategies tailored for LCLMs and conduct extensive experiments on various popular models. Empirical results demonstrate that our attacks can accurately infer membership status in most cases, e.g., 90.66\% attack F1-score on Multi-document QA datasets with LongChat-7b-v1.5-32k, highlighting significant risks of membership leakage within LCLMs’ input contexts. Furthermore, we examine the underlying reasons why LCLMs are susceptible to revealing such membership information.
\end{abstract}

\section{Introduction}
Long-Context Language Models (LCLMs)~\cite{abdin2024phi, young2024yi, liu2024world, team2024gemini} have recently garnered significant attention for enabling large language models (LLMs) to understand and process longer sequences. Recent systems equipped with LCLMs can automatically search massive external data and incorporate it into the contexts of these models, thus making faithful predictions and mitigating hallucinations and knowledge staleness. Existing studies on LCLMs primarily concentrate on addressing the \textit{lost-in-the-middle} problem~\cite{shi2023large, li2024long, liu2024lost} or accelerating their inference speed~\cite{zhang2023h2o, li2024snapkv, shi2024discovering}, leaving the privacy risks largely unexplored.
However, the integration of massive retrieved data poses significant privacy risks on the input context of LCLMs, which may contain highly sensitive data such as medical records or personal identities.


In this work, we focus on a fundamental attack known as Membership Inference Attacks (MIAs)~\cite{mattern2023neighbor, shi2023mink, ko2023practical} to evaluate and quantify the privacy and security of the LCLM’s input context. The goal of MIAs against LCLMs is to determine whether a given document or sequence (i.e., the target sample) is included in the input context of the given LCLM. Successful MIAs can lead to severe privacy risks for the data providers of LCLMs. For example, if a target sample in a medical domain LCLM’s database is identified, it may reveal the disease history of the data provider.

However, existing MIAs are typically tailored for traditional classification or generative machine learning models. These approaches rely on the model’s tendency to overfit its training samples, which results in distinguishable predictions for the training samples. In contrast, LCLMs do not train the LLMs on their context and lack the overfitting characteristic on the samples within the external inputs. Moreover, existing MIA methods targeting LLMs~\cite{mattern2023neighbor, shi2023mink} primarily focus on sentence-level. The documents within the context of LCLMs typically contain thousands of tokens, which further reduces the overfitting tendency.

In this paper, we propose six MIA methods to quantify the membership leakage for LCLMs leveraging the generation loss or semantic similarity between the target document and generated continuations. Our basic idea is that if a sample lies in the input context of a LCLM, when we utilize relevant contents to query the model, it tends to retrieve and reasoning over this document and generate contents related to the target sample. These associations can be detected by measuring the model’s perplexity or the semantic similarity between its outputs and the target document. To amplify these signals, we propose to prompt LCLMs with only small snippets from the target documents, which strongly activates the retrieval power of these models, causing them to omit crucial membership information from their generation loss or semantic similarity.



This paper makes the following contributions:
\begin{itemize}
	\item We present the first MIA against LCLMs, which reveals the membership privacy risks of the long-context within current LCLMs. We propose six types MIAs including both probability-based and text-only attacks, which addresses the significant gap in understanding and quantifying the membership leakage within LCLM's long context.
	\item We find that even though LCLMs cannot fully utilize the information contained in the context, they indeed know whether certain documents or sequences are included, and such signal can be captured from model responses. Based on this observation, we show for the first time that the generation loss or semantic similarity may leak the membership information and can be leveraged to develop MIAs against LCLMs.
	\item We evaluate the performance of six proposed attacks on several popular LCLMs, and experiment results demonstrate that our attacks can achieve a strong performance compared with existing MIAs, even under strict text-only attack scenarios.
\end{itemize}

\begin{figure}[h]
	\begin{minipage}{0.34\linewidth}
		\vspace{3pt}
		\centerline{\includegraphics[width=\textwidth]{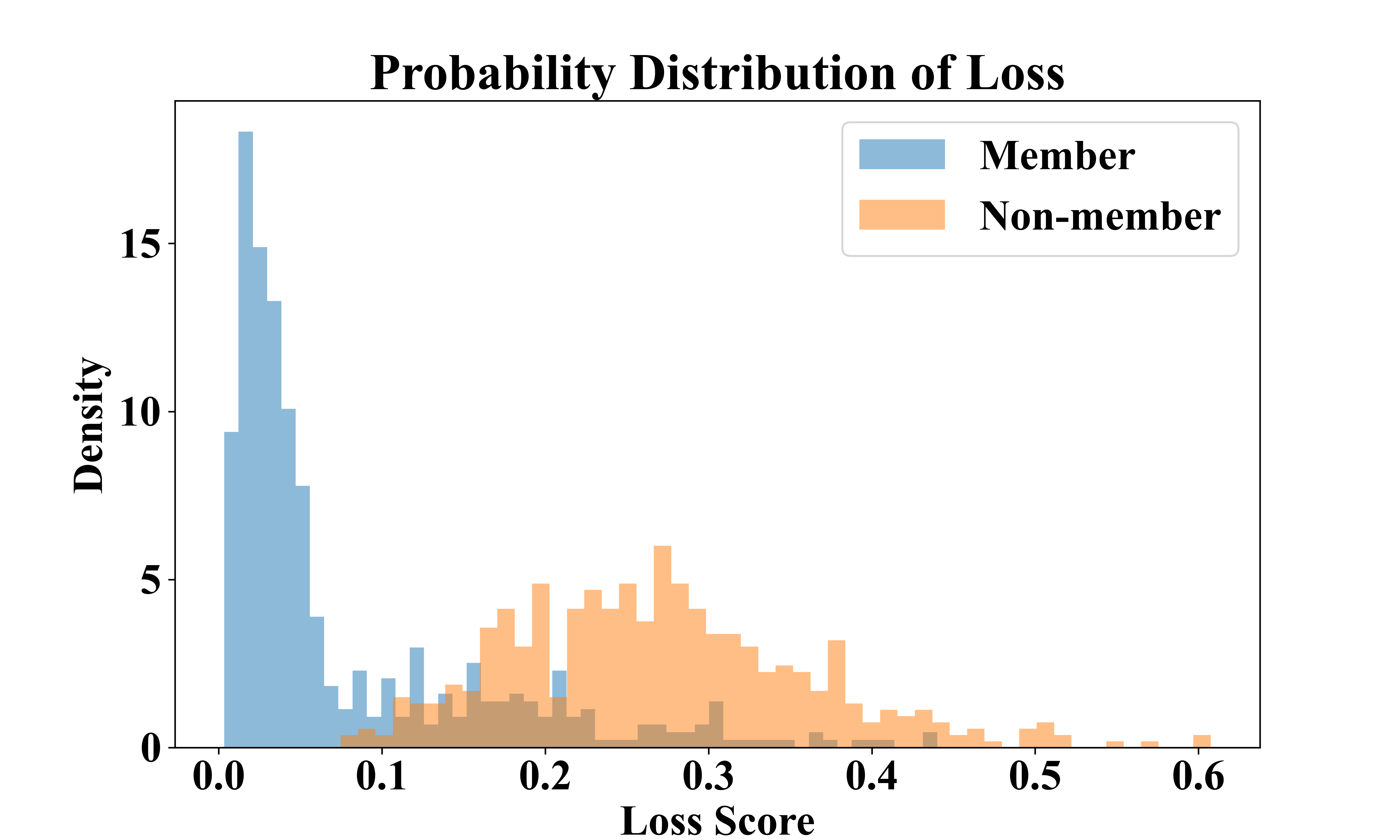}}
		\centerline{\footnotesize (1) Loss}
	\end{minipage}
	\begin{minipage}{0.34\linewidth}
		\vspace{3pt}
		\centerline{\includegraphics[width=\textwidth]{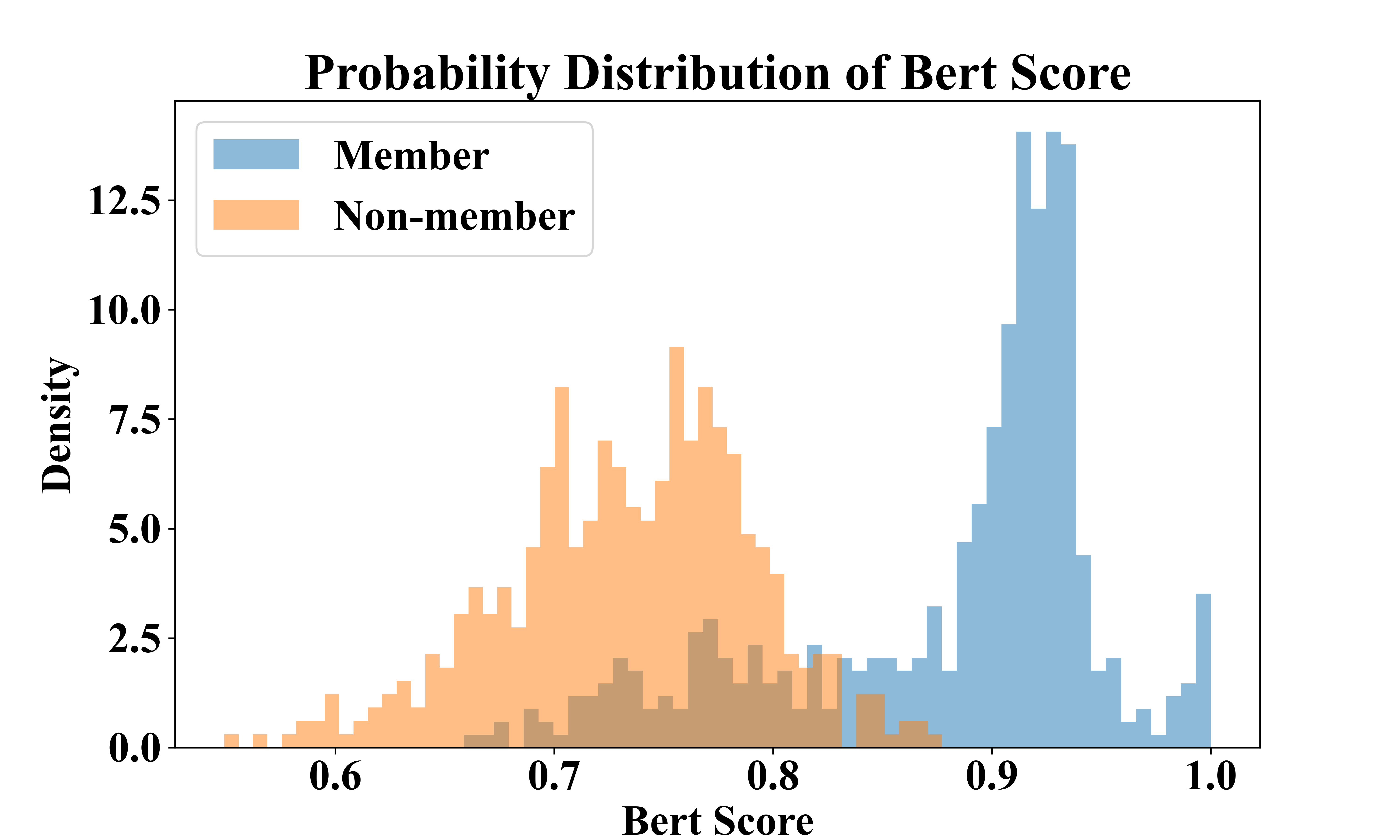}}
		
		\centerline{\footnotesize (2) Bert Score}
	\end{minipage}
	\begin{minipage}{0.34\linewidth}
		\vspace{3pt}
		\centerline{\includegraphics[width=\textwidth]{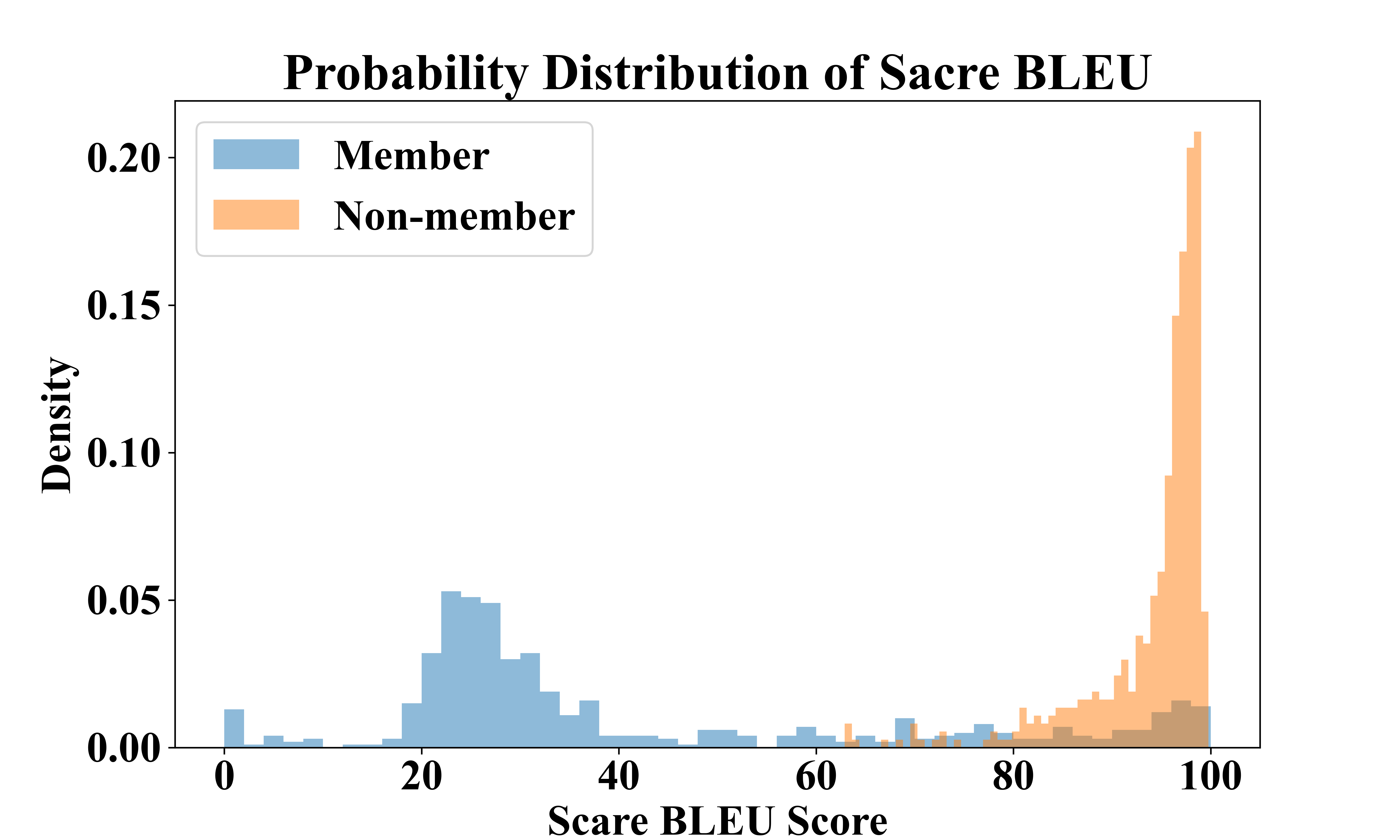}}
		
		\centerline{\footnotesize (3) N-Sacre BLEU Score}
	\end{minipage}
	
	\caption{Density distributions of members v.s. non-members on several metrics. We leverage LongChat-7b-v1.5-32k and $1,000$ samples ($500$ members and $500$ non-members) from NaturalQuestion multi-document QA dataset (30 documents) to plot these figures. Note that we set the gold document to be the 15th document, which significantly improves the difficulty for inference attacks against LCLMs~\cite{liu2024lost}.}
	\label{fig4}
\end{figure}

\section{Preliminary}
\subsection{Long-Context Language Modeling}
Oringial LLMs, such as Llama~\cite{touvron2023llama} with a context length of $2048$ and Llama2~\cite{touvron2024llama} with $4096$, are typically pre-trained using a pre-defined context window length. However, training LLMs with longer context size from scratch is time-consuming and extensive.
Therefore, there emerge increasing research efforts in developing strategies to extend LLMs' context window size. 
Some works augment LLMs via fetching related documents and compress them into the contexts~\cite{karpukhin2020dense}.
Recent studies primarily focus on adjusting model attention mechanism~\cite{meister2021sparse, ding2023longnet} or developing novel position encoding approaches~\cite{press2021train, ding2024longrope}. Among them, sparse attention~\cite{meister2021sparse} and dilated attention~\cite{ding2023longnet} aim to improve the efficiency of context scaling, while ALiBi~\cite{press2021train} and RoPE~\cite{ding2024longrope} improve the extrapolation in Transformer architectures~\cite{vaswani2017attention}.
\subsection{Membership Inference Attacks against LLMs}
Membership inference attack stands as one of the most fundamental privacy risks in machine learning models, where the attacker aims to determine whether a given sample belongs to the training set of the target machine learning models. Such attacks have been extensively studied in the realm of LLMs, where adversaries develop specifically designed prompts for querying these models and breach private information based on their responses. These MIAs have been applied to mask language models~\cite{mireshghallah2022quantifying}, decoder-only language models~\cite{mattern2023neighbor, shi2023mink}, RAG systems~\cite{li2024seeing, anderson2024my}, and LLM in-context learning~\cite{wen2024membership}.

The fundamental insight of existing works are typically based on the distinct behaviors of LLMs when processing training and testing samples. These studies often necessitate the model's generation probabilities, for measuring the generation coherence or perplexity of LLMs. Note that in this work, we explore the both scenarios where probability is available or not. In addition, we conduct the first investigation on the membership privacy risks within the external contexts of LCLMs.
\subsection{Threat Model}
\textbf{Attacker's Goal:} Given a target document $x_t$ and a victim LCLM $M$, the attacker's goal is to infer whether $x_t$ is included in the input long-context $C$ or not:
\begin{equation}
	\mathcal{A}(x_t \vert M, C) \rightarrow \textbf{Member}/\textbf{Non-member}
\end{equation}

\textbf{Attacker's Knowledge:} The attacker does not have access to the LCLM's parameters, the inference hidden states, nor any information about the prepending long-context.

\textbf{Attacker's Capacity:} As in previous works~\cite{shi2023mink, wen2024membership}, the attacker in our paper can query the victim LCLM and obtain the output text along with the prediction probabilities.

\subsection{System Prompt}
Following existing works~\cite{liu2024lost, gao2024insights}, we utilize the system prompt with the format presented below:

\begin{tcolorbox}[
	colframe=black!75,          
	colback=gray!10,            
	coltitle=black,             
	sharp corners,              
	width=\linewidth,           
	boxrule=0.5mm,              
	fontupper=\fontsize{8pt}{8pt}\selectfont, 
	boxsep=1mm,                 
	left=1mm,                   
	right=1mm,                  
	top=0.6mm,                    
	bottom=0.6mm,                  
	title = {\textcolor{white}{System Prompt}}
	]
	\label{systemprompt}
	Write a high-quality answer for the given question using only the provided search results (some of which might be irrelevant). \\
	Document [0] (Title: Nobel Prize in Physics) receive a diploma, a medal and a document confirming $\ldots$ \\
	Document [1] (Title: Norwegian Americans) science, Ernest Lawrence won the Nobel Prize in Physics in $\ldots$ \\
	Document [2] (Title: Nobel Prize in Physics) Nobel Prize in Physics The Nobel Prize in Physics $\ldots$ \\
	Document [3] (Title: Ecole normale superieure (Paris)) was also awarded the Abel prize. In addition $\ldots$ \\
	Document [4] (Title: List of Nobel laureates in Physics) The first Nobel Prize in Physics was awarded in 1901 $\ldots$ \\
	Document [5] (Title: Nobel Prize in Physics) rendered by the discovery of the remarkable rays (or x-rays) $\ldots$ \\
	Document [6] (Title: E. C. George Sudarshan) had developed the breakthrough. In 2007, Sudarshan told the $\ldots$ \\
	Document [7] (Title: Svante Arrhenius) Wilhelm Ostwald, Theodore Richards) and to attempt to deny them to $\ldots$ \\
	Document [8] (Title: Scientist) and pursued through a unique method, was essentially in place $\ldots$ \\
	Document [9] (Title: Nobel Prize in Chemistry) on December 10, the anniversary of Nobel's death. The first $\ldots$ \\
	\{User Question: who got the first nobel prize in physics?\} \\
	\{\textbf{Here adversaries can input their attack prompts.}\} \\
	\{Answer:\}
\end{tcolorbox}

\section{Our Method}
We propose to conduct MIA against LCLMs with the following approaches, which can be categorized into two types: \textit{probability-based attacks} and \textit{text-only attacks}.

When the adversaries can get the generation probabilities from open API, they can launch attacks based on these logits, which includes:
\begin{itemize}
	\item \textbf{Logits Attack.} This method adds the following attack prompt to the system prompt after the user question, and just asks the victim model to predict yes or no. Then it manually check and compare the generation probabilities of these tokens: (yes, Yes, \& YES) and (no, No, \& NO).
	
	\begin{tcolorbox}[
		colframe=black!75,          
		colback=gray!10,            
		coltitle=black,             
		sharp corners,              
		width=\linewidth,           
		boxrule=0.5mm,              
		fontupper=\fontsize{8pt}{8pt}\selectfont, 
		boxsep=1mm,                 
		left=1mm,                   
		right=1mm,                  
		top=0.6mm,                    
		bottom=0.6mm,                  
		title = {\textcolor{white}{Logits Attck Prompt}}
		]
		\{User Question: {\textit{question}}\} \\
		\{Notice: Ignore the question and answer me with just yes or no. Have you seen this verbatim sentence from the all documents: \{$x_t$\}? If you have seen it before, please be confident and say yes. If you do have not seen it in your documents or you are not sure, please be honest and just answer no.\} \\
		\{Answer:\}
	\end{tcolorbox}
	
	\item \textbf{Loss Attack.} This method first divide the target document $x_t$ into prefix $x_p$ and suffix $x_s$, and then adds the following attack prompt to the system prompt after user question. This instruction asks the victim model to complete the given prefix. Then this method measure the generation loss values only on the continuations which shares the same token length as $x_s$.
	
	\begin{tcolorbox}[
		colframe=black!75,          
		colback=gray!10,            
		coltitle=black,             
		sharp corners,              
		width=\linewidth,           
		boxrule=0.5mm,              
		fontupper=\fontsize{8pt}{8pt}\selectfont, 
		boxsep=1mm,                 
		left=1mm,                   
		right=1mm,                  
		top=0.6mm,                    
		bottom=0.6mm,                  
		title = {\textcolor{white}{Loss Attack Prompt}}
		]
		\{User Question: {\textit{question}}\} \\
		\{Notice: Ignore the question and just focus on my request now below:\} \\
		\{Please complete this sentence based on the information collected from the documents above: \{$x_p$\} \} \\
		\{Answer:\}
	\end{tcolorbox}
	
\end{itemize}

Under more realistic and difficult scenarios, the adversaries can only query the victim LCLM and get its final predicted texts. Thus, we propose the following three text-only attack approaches:
\begin{itemize}
	\item \textbf{Inquiry Attack.}~\cite{anderson2024my, wen2024membership} This method directly asks the model using the following attack prompt after the user question. Different from the Logits Attack, this methods only receives the answer of one single token (yes or no).
	
	\begin{tcolorbox}[
		colframe=black!75,          
		colback=gray!10,            
		coltitle=black,             
		sharp corners,              
		width=\linewidth,           
		boxrule=0.5mm,              
		fontupper=\fontsize{8pt}{8pt}\selectfont, 
		boxsep=1mm,                 
		left=1mm,                   
		right=1mm,                  
		top=0.6mm,                    
		bottom=0.6mm,                  
		title = {\textcolor{white}{Inquiry Attack Prompt}}
		]
		\{User Question: {\textit{question}}\} \\
		\{Notice: Ignore the question and answer me with just yes or no. Have you seen this verbatim sentence from the all documents: \{$x_t$\}? If you have seen it before, please be confident and say yes. If you do have not seen it in your documents or you are not sure, please be honest and just answer no.\} \\
		\{Answer:\}
	\end{tcolorbox}
	
	\item \textbf{Bert Score Attack.} This method is similar to the Loss attack, but it measures the Bert Score~\cite{zhang2019bertscore} between gold suffix and the continuation generated by the victim model.
	\item \textbf{Sacre BLEU Attack.} This method is similar to the Loss attack, but it measures the Scare BLEU Score~\cite{post2018call} between gold suffix and the continuation generated by the victim model.
\end{itemize}

Based on the above introduced methods, we develop a strong attack strategy which necessitate the access to the generation probabilities:

\begin{itemize}
	\item \textbf{Meta-Classifier.} Based on Loss attack, we propose to divide the target document with different token length for multiple times, each time we query the victim model and collect the loss, Bert score, and Sacre BLEU score. Finally we combine all these values into a vector, and leverage all the vectors as membership features. Finally we conduct MIA by training the attack model using these membership features.
\end{itemize}

\begin{table*}[htbp!]
	\caption{MIA performance of six proposed methods against the victim model of LongChat-7b-v1.5-32k (The highest values for each dataset are highlighted in bold). The dataset is Natural Question multi-document QA~\cite{liu2024lost}. The position of MIA target document is always right in the middle of all documents (5th, 10th, and 15th).}
	\centering
	\label{maintable}
	\renewcommand{\arraystretch}{0.9}
	\fontsize{8pt}{12pt}\selectfont
	\arrayrulecolor[gray]{0.7}
	\resizebox{\textwidth}{!}{%
		\begin{tabular}{l|cccc|cccc|cccc}
			\Xhline{3\arrayrulewidth}
			{\raisebox{-0.15em}{\textbf{Task}}} & \multicolumn{4}{c|}{\raisebox{-0.15em}{10-Document QA}} & \multicolumn{4}{c|}{\raisebox{-0.15em}{20-Document QA}} & \multicolumn{4}{c}{\raisebox{-0.15em}{30-Document QA}}\\
			\hline
			\arrayrulecolor{black}
			\raisebox{-0.15em}{\textbf{Metric}} & \raisebox{-0.15em}{Accrucy} & \raisebox{-0.15em}{Precision} & \raisebox{-0.15em}{Recall} & \raisebox{-0.15em}{F1-Score} & \raisebox{-0.15em}{Accuracy} & \raisebox{-0.15em}{Precision} & \raisebox{-0.15em}{Recall} & \raisebox{-0.15em}{F1-Score} & \raisebox{-0.15em}{Accrucy} & \raisebox{-0.15em}{Precision} & \raisebox{-0.15em}{Recall} & \raisebox{-0.15em}{F1-Score} \\
			\hline \hline
			\rowcolor{gray!20}
			{\raisebox{-0.15em}{\textbf{Scenario}}} & \multicolumn{12}{c}{\raisebox{-0.15em}{\textbf{Probability-based Attack}}}\\
			\arrayrulecolor[gray]{0.7}\hline
			\arrayrulecolor{black}
			\raisebox{-0.15em}{Logits Attack} & \raisebox{-0.15em}{70.50} & \raisebox{-0.15em}{62.93} & \raisebox{-0.15em}{\textbf{99.80}} & \raisebox{-0.15em}{77.18} & \raisebox{-0.15em}{68.20} & \raisebox{-0.15em}{61.18} & \raisebox{-0.15em}{\textbf{99.60}} & \raisebox{-0.15em}{75.79} & \raisebox{-0.15em}{65.30} & \raisebox{-0.15em}{59.27} & \raisebox{-0.15em}{\textbf{97.80}} & \raisebox{-0.15em}{73.81} \\
			\raisebox{-0.15em}{Loss Attack} & \raisebox{-0.15em}{88.90} & \raisebox{-0.15em}{92.75} & \raisebox{-0.15em}{84.40} & \raisebox{-0.15em}{88.38} & \raisebox{-0.15em}{89.10} & \raisebox{-0.15em}{91.33} & \raisebox{-0.15em}{86.40} & \raisebox{-0.15em}{88.79} & \raisebox{-0.15em}{87.30} & \raisebox{-0.15em}{\textbf{98.44}} & \raisebox{-0.15em}{75.80} & \raisebox{-0.15em}{85.65} \\
			\raisebox{-0.15em}{Meta Classifier} & \raisebox{-0.15em}{\textbf{92.40}} & \raisebox{-0.15em}{96.91} & \raisebox{-0.15em}{87.60} & \raisebox{-0.15em}{\textbf{92.01}} & \raisebox{-0.15em}{\textbf{91.90}} & \raisebox{-0.15em}{\textbf{92.49}} & \raisebox{-0.15em}{91.20} & \raisebox{-0.15em}{\textbf{91.84}} & \raisebox{-0.15em}{\textbf{91.10}} & \raisebox{-0.15em}{95.36} & \raisebox{-0.15em}{86.40} & \raisebox{-0.15em}{\textbf{90.66}} \\
			\hline \hline
			\rowcolor{gray!20}
			{\raisebox{-0.15em}{\textbf{Scenario}}} & \multicolumn{12}{c}{\raisebox{-0.15em}{\textbf{Text-Only Attack}}}\\
			\arrayrulecolor[gray]{0.7}\hline
			\arrayrulecolor{black}
			\raisebox{-0.15em}{Inquiry Attack} & \raisebox{-0.15em}{69.20} & \raisebox{-0.15em}{62.09} & \raisebox{-0.15em}{98.60} & \raisebox{-0.15em}{76.19} & \raisebox{-0.15em}{66.90} & \raisebox{-0.15em}{60.49} & \raisebox{-0.15em}{97.40} & \raisebox{-0.15em}{74.63} & \raisebox{-0.15em}{63.30} & \raisebox{-0.15em}{57.91} & \raisebox{-0.15em}{97.40} & \raisebox{-0.15em}{72.63} \\
			\raisebox{-0.15em}{Bert Score Attack} & \raisebox{-0.15em}{90.10} & \raisebox{-0.15em}{94.85} & \raisebox{-0.15em}{84.80} & \raisebox{-0.15em}{89.55} & \raisebox{-0.15em}{90.50} & \raisebox{-0.15em}{95.31} & \raisebox{-0.15em}{85.20} & \raisebox{-0.15em}{89.97} & \raisebox{-0.15em}{88.80} & \raisebox{-0.15em}{89.27} & \raisebox{-0.15em}{88.20} & \raisebox{-0.15em}{88.73} \\
			\raisebox{-0.15em}{Scare BLEU Attack} & \raisebox{-0.15em}{90.60} & \raisebox{-0.15em}{\textbf{96.99}} & \raisebox{-0.15em}{83.80} & \raisebox{-0.15em}{89.91} & \raisebox{-0.15em}{91.10} & \raisebox{-0.15em}{96.81} & \raisebox{-0.15em}{85.00} & \raisebox{-0.15em}{90.52} & \raisebox{-0.15em}{90.70} & \raisebox{-0.15em}{95.73} & \raisebox{-0.15em}{85.20} & \raisebox{-0.15em}{90.16} \\
			\Xhline{3\arrayrulewidth}
		\end{tabular}
	}
	\vspace{-5pt}
\end{table*}

\section{Experiment Evaluation}
\subsection{Experiment Setting}
\textbf{Datasets.} We evaluate our attacks on \textbf{Multi-Document QA}~\cite{liu2024lost} and \textbf{KV-retrieval}~\cite{liu2024lost} with gold document at different position in the context. We also evaluate the performance on general long-context benchmark from \textbf{LongBench}~\cite{bai2023longbench}, which includes multi-document QA, single-document QA, summarization, few-shot learning, synthetic tasks, and
code completion, totally 16 tasks with an average length of 37k tokens.

\textbf{LCLMs.} We evaluate our methods on a wide range of state-of-the-art open source LLMs, including: LongChat-7b-v1.5-64k~\cite{li2023long}, LongChat-13b-16k~\cite{li2023long}, Vicuna-13b-v1.5-16k~\cite{chiang2023vicuna}, Mistral-7b-Instruct-v0.2~\cite{jiang2023mistral}, Gemma-1.1-7b-it~\cite{team2024gemma}, and Qwen1.5-7b-chat.

\textbf{Attack Setting.} When evaluating on each dataset, we combine all documents into the long context and regard them as members at different locations. The non-members come from documents in another dataset. For each dataset, we randomly select $500$ members and non-members to construct the reference dataset, and additional $1,000$ samples are utilized as the test dataset for accessing the performance of all MIA methods.
In the Loss Attack, we by default divide the target document into four equal snippets, and utilize the first snippet for inserting after the user question. Then, we calculate the generation loss for the next three snippets.

\textbf{Evaluation Metrics.} Following classic MIA methods~\cite{shokri2017membership}, we leverage \emph{accuracy}, \emph{precision}, \emph{recall}, and \emph{F1-score} to evaluate all proposed methods. Specifically, \emph{accuracy} represents the proportion of target samples whose membership are correctly inferred. \emph{Precision} is the ratio of predicted members that are indeed members. \emph{Recall} presents the fraction of training records which are correctly inferred as members. To provide a more intuitive and precise evaluation, we also report their \emph{F1-scores}.

\textbf{Details for Meta-Classifier.} We first divide the target document $x_t$ into 2, 4, 6, 8, and 10 pieces with equal token length. Then we always regard the first one piece as the prefix and collect the loss, Bert Score, BLEU score based on the above attack methods. Subsequently, we utilize totally $15$ values and combine them into a vector, which is regarded as the membership feature for this document. We train a fully connected layer as the attack model with all the membership features from $1,000$ reference samples, and apply inference attack on $1,000$ testing samples.
\subsection{Evaluation Results}
\subsubsection{Main Results}
We evaluate the performance of six proposed attack strategies on the LongChat-7b-v1.5-32k model using Multi-Document QA datasets with varying numbers of documents, as detailed in Table~\ref{maintable}. Our findings reveal that both the Bert Score Attack and the Sacre BLEU Attack achieve high inference effectiveness, with attack F1-scores exceeding 85\% across all three datasets, even in text-only scenarios. This supports our observation that LCLMs, when prompted with sequences within their input context, tend to retrieve relevant documents meticulously and generate highly pertinent outputs.
Notably, the Meta-Classifier outperforms all other methods, achieving attack accuracy scores above 90\% across all datasets. This highlights the advantage of considering both semantic similarity and language coherence in attack strategies. These results underscore the substantial risks of membership leakage within LCLM contexts—a concern that has been largely unexplored and underestimated in prior research.

In contrast, the Logits Attack and Inquiry Attack yield lower effectiveness, with accuracy scores below 70\%. We hypothesize that utilizing the entire documents and instructing them to predict a single token is less effective, as it does not fully leverage the model’s strong retrieval and reasoning capabilities. Additionally, asking the LCLM to respond with a simple yes or no introduces position bias, as noted in~\cite{wang2024eliminating}.

\section{Conclusion}

In this paper, we present the first MIAs against LCLMs, which reveals the membership privacy risks of the long-context within current LCLMs. We propose six types MIAs including both probability-based and text-only attacks, which addresses the significant gap in understanding and quantifying the membership leakage within LCLM's long context. We demonstrate that the semantic similarity and generation loss differences between member and non-member samples within the LCLM's context, relative to their corresponding generated contents, can substantially undermine membership privacy. 
Extensive experimental results indicate that existing LCLMs confront severe MIA risks with respect to their long-contexts.
This work represents a significant step towards understanding the privacy leakage risks associated with LCLMs and illuminates the path for developing more robust defense mechanisms against MIAs in these models.

\newpage
\bibliographystyle{IEEEtran}
\bibliography{template}
\end{document}